\documentclass{article}

\usepackage{microtype}
\usepackage{graphicx}
\usepackage{subfigure}
\usepackage{booktabs} 

\usepackage[table,xcdraw]{xcolor}

\usepackage{hhline}
\usepackage{array, makecell}
\usepackage{graphicx}
\usepackage{epstopdf}
\usepackage{tabularx}

\usepackage{hyperref}

\usepackage{amsmath}

\usepackage[accepted]{icml2019}

\icmltitlerunning{Power Grid Cascading Failure Mitigation by Reinforcement Learning}

\def\hlinewd#1{%
\noalign{\ifnum0=`}\fi\hrule \@height #1 %
\futurelet\reserved@a\@xhline}

\begin{document}

\twocolumn[
\icmltitle{Power Grid Cascading Failure Mitigation by Reinforcement Learning}



\icmlsetsymbol{equal}{*}

\begin{icmlauthorlist}
\icmlauthor{Yongli Zhu}{equal,tamu}
\end{icmlauthorlist}

\icmlaffiliation{tamu}{ECE Dept., Texas A\&M University, College Station, USA}

\icmlcorrespondingauthor{Yongli Zhu}{zylpascal@gmail.com, yongliz@tamu.edu}
\icmlkeywords{Machine Learning, ICML}

\vskip 0.3in
]



\printAffiliationsAndNotice{}  

\begin{abstract}
This paper proposes a cascading failure mitigation strategy based on Reinforcement Learning (RL). The motivation of the Multi-Stage Cascading Failure (MSCF) problem and its connection with the challenge of climate change are introduced. The bottom-level corrective control of the MCSF problem is formulated based on DCOPF (Direct Current Optimal Power Flow). Then, to mitigate the MSCF issue by a high-level RL-based strategy, physics-informed reward, action, and state are devised. Besides, both shallow and deep neural network architectures are tested. Experiments on the IEEE 118-bus system by the proposed mitigation strategy demonstrate a promising performance in reducing system collapses.
\end{abstract}

\section{Introduction}
\label{submission}

Increasing renewable sources (e.g., wind, solar) are integrated into modern power grids to reduce emissions. However, due to the intermittent natures of those renewable sources, the original power grid can become fragile, i.e., more easily affected by various risks. Among those risks, the cascading failure is one of the most challenging issues necessary to be addressed \cite{SunBook}. A \textit{cascading failure} is defined as a series of consecutive malfunctions of physical components (e.g., power transmission lines, power substations). A cascading failure is typically caused by an unexpected natural disaster such as a hurricane, typhoon, or flood. Severe cascading failures can lead to a total system collapse (i.e., disintegrated into small energized pieces) or even a blackout event (i.e., loss of electricity for the entire city or country). To recover the power grid to a healthy state, backup generators need to be turned on. The relevance of the cascading failure problem to climate change is: 

1) Given the increase in extreme weather events due to climate change, having a stable power grid is critical for renewable resources integration; 

2) The backup generators are typically fossil-fuel (e.g., coal, diesel) based units that emit greenhouse gases. 

Therefore, it is meaningful to develop nuanced strategies to mitigate a cascading failure at its early stage with as little generation cost as possible.

The cascading failure mitigation can be regarded as a stochastic dynamic programming problem with unknown information about the risk of failures. Previous researches try to tackle this problem based on either mathematical programming methods or heuristic methods. For example, bi-level programming is used to mitigate cascading failures when energy storages exist \cite{Du1st}. In \cite{HAN2018}, an algorithm based on percolation theory is employed for mitigating cascading failures by using UPFC (Unified Power Flow Controller) to redistribute the system power flow more evenly. In \cite{Tootaghaj18}, a recovery plan for load-serving from cascading failures was put forward considering the uncertainty of failure locations. In \cite{Garcia18}, a cascade control algorithm using load shedding considering the communication network delays for power grids was proposed to reduce the failure of power lines. In \cite{Shuvro17}, By characterizing the cascading-failure dynamics as a Markov chain model, it is found that the impact of human-operator action will have a significant impact on cascading failures. 

However, some of the above research share the same limitation: unrealistic assumptions are often made,  which yield impractical control strategies in terms of time or economic cost. Different from communication networks and social networks, the power grid is not a pure-mathematical \textit{graph}, but a physical grid.  Its node (called ``bus” in power system terminology) and edge (called ``branch'' in power system terminology) are both physical entities that can not be added or removed arbitrarily. On the other hand, most power system lines are equipped with automatic protection-relay devices, which can trip the line when the line current/power/temperature exceeds certain thresholds in a pre-defined time window.  Thus, in this paper, the main focus is on branch failures rather than node failures. 

Meanwhile, some emerging artificial intelligence technologies, such as reinforcement learning (RL) and deep learning (DL), have nourished both the fields of power systems and control theory \cite{Kiumarsi18}. In \cite{Glavic17}, a holistic literature review is given on the recent development of RL applications in the power system area, though topics regarding cascading failure are not covered. In \cite{Vlachogiannis04}, the RL method is used for reactive power control. In \cite{LIU18}, voltage restoration for an islanded microgrid is achieved via a distributed RL method. In \cite{ZHU18}, an application for disturbance classification is proposed based on image embedding and convolutional neural network (CNN). In \cite{YAN18}, deep learning is applied in power consumption forecasting. However, the application of RL or DL for cascading failure study are less reported.  

In this paper, a reinforcement learning approach is designed for the mitigation of cascading failures with the following contributions:

\begin{itemize}
\item 1)	Propose and formulate a novel problem called \textit{Multi-Stage Cascading Failure} (MSCF) for the first time.
\item 2)	Present a systematic reinforcement learning framework to tackle the MSCF problem. Similar to AlphaGo \cite{Silver16}, a ``two-player game“ idea is utilized in this study.  
\item 3)	Unlike some previous study which treats power grid purely as a graph model with no or less physical background, this paper uses a professional power system simulator as the \textit{environment} in the RL framework to better reflect actual characteristics of the power system. In this way,  the learning result is more convincing, and the trained mitigation strategy will be more practical.
\end{itemize}

The remaining parts of this paper are organized as follows. Section 2 proposes an RL-based control framework for the mitigation of cascading failures. Section 3 presents the case study and results analysis. Finally, conclusions and future directions are given in Section 4.

\section{Multi-Stage Cascading Failure Control}

\subsection{Multi-Stage Cascading Failure (MSCF) Problem}

Firstly, the following definitions are given:

\textit{Generation}: one event of the cascading failures within one stage, e.g., a line tripping \cite{Qi2015TPWRS}. 

\textit{Stage}: after an attack (e.g., one line is broken by a natural disaster), the grid \textit{evolves} with a series of potential \textit{generations} (e.g., line tripping events if the line thermal limits are reached). At the end of each stage, the power system will either 1) reach a new equilibrium point if the ACPF (Alternative Current Power Flow) converges and all branch flows are within secure limits or 2) become collapsed.

In conventional cascading failure analysis, typically only one stage is considered \cite{Qi2017TWPRS}. However, in certain situations, succeeding stages might follow shortly. For example, a wind storm results in one generation, in which certain lines are lost, and the system reaches a new steady state. Then, shortly, a new stage is invoked by tripping an important line due to the misoperation of human-operator or relay protection. As an example, Table \ref{table-1} and \ref{table-2} list the simulation results of the IEEE 118 system for a two-stage MSCF problem in two independent episodes. 

\begin{table}[b]
\caption{Result of Episode-1.}
\label{table-1}
\vskip 0.1in
\begin{center}
\begin{small}
\begin{sc}
\begin{tabular}{m{1.5cm}lccl}
\toprule
\textit{\textbf{Stage}}-1
    &                       &   \makecell{ACPF \\ Converge}
                                        &   \makecell{Over-limit \\ Lines} \\
\cline{1-1}
    & \textit{Generation}-1 & Yes  & 0 \\ 
\hline
\textit{\textbf{Stage}}-2
    &                       &   \makecell{ACPF \\ Converge}
                                        &   \makecell{Over-limit \\ Lines} \\
\cline{1-1}
    & \textit{Generation}-2 & Yes  & 0 \\ 
\hhline{====} \\
\textit{\textbf{Result}}
	&        &    Win  & \\
\bottomrule
\end{tabular}
\end{sc}
\end{small}
\end{center}
\vskip -0.1in
\end{table}

\begin{table}[b]
\caption{Result of Episode-2.}
\label{table-2}
\vskip 0.1in
\begin{center}
\begin{small}
\begin{sc}
\begin{tabular}{m{1.5cm}lccl}
\toprule
\textit{\textbf{Stage}}-1
    &                       &   \makecell{ACPF \\ Converge}
                                        &   \makecell{Over-limit \\ Lines} \\
\cline{1-1}
    & \textit{Generation}-1 & Yes  & 2 \\ 
    & \textit{Generation}-2 & Yes  & 0 \\ 
\hline
\textit{\textbf{Stage}}-2
    &                       &   \makecell{ACPF \\ Converge}
                                        &   \makecell{Over-limit \\ Lines} \\
\cline{1-1}
    & \textit{Generation}-1 & Yes  & 4 \\ 
    & \textit{Generation}-2 & Yes  & 2 \\ 
    & \textit{Generation}-3 & Yes  & 2 \\ 
    & \textit{Generation}-4 & Yes  & 3 \\ 
    & \textit{Generation}-5 & Yes  & 10 \\ 
    & \textit{Generation}-6 & Yes  & 20 \\ 
    & \textit{Generation}-7 & No  & -- \\
\hhline{====} \\
\textit{\textbf{Result}}
	&        &    Lose  & \\
\bottomrule
\end{tabular}
\end{sc}
\end{small}
\end{center}
\vskip -0.1in
\end{table}    

A na\"ive way to handle this complicated multi-stage scenario is to tackle each stage independently by existing method  e.g. SCOPF (Security Constrained Optimal Power Flow). However, merely using SCOPF may not work well due to the overlook of the correlations between any two consecutive stages. For example, the previous study \cite{Zhu2014,Chen2007} found that sequentially attacking each line one by one can sometimes achieve a more severe effect (i.e., more components loss) than attack multiple lines simultaneously. Thus, the MSCF problem should be considered from a \textit{holistic} perspective.

\subsection{Mimicking the corrective controls by DCOPF}

When a failure event happens, the following DCOPF (Direct Current Optimal Power Flow) is adopted \cite{Changshen2019gAccess} to mimic the bottom-level control measures, i.e., changing generator outputs and shedding loads (if necessary).

\begin{equation}
\begin{aligned}
\min_{p_i,p_j} \quad & \sum_{i \in G}c_i p_i + \sum_{j \in D}d_j(p_j-P_{dj})\\
\textrm{s.t.} \quad & ~\text{\textbf{F}} = \text{\textbf{Ap}}\\
  &\sum_{k=1}^{n}p_k=0   \\
  & P_{dj}  \leq  p_j \leq 0, &j \in D   \\
  & P_{gi}^{min}  \leq  p_j \leq P_{gi}^{max}, &i \in G   \\
  & -F_{L}^{max}  \leq  F_l \leq F_{L}^{max},&l \in L   \\
\end{aligned}
\end{equation}

Where, \textit{n} is the total bus number. \textit{G}, \textit{D}, \textit{L} are respectively the generator set, load set and branch set; \textbf{F} = $[F_l] (l \in L)$ represents the branch flow; $p_i (i \in G)$ is the generation dispatch for the \textit{i}-th generator; $p_j (j \in D$) is the load dispatch for the \textit{j}-th load; \textbf{p} = $[p_k]$, $k={1...n}$ represents the (net) nodal power injections. \textbf{A} is a constant matrix to associate the (net) nodal power injections with the branch flows. $P_{dj}$ is the normal demand value for the \textit{j}-th load; $c_i$ is the given cost coefficient of generation; $d_i$ is the given cost coefficient of load shedding. $p_i (i \in G)$ and $p_j (j \in D)$ are the decision variables for generators and load respectively.

\subsection{Apply RL for MSCF problem}

To apply RL to a specific power system problem, the first step is to map physical quantities of the power grid to components of the RL framework, i.e., reward, action, and state.

1) Reward design (of each stage)\\
$\bullet$ $-$Total generation cost (i.e., the negative objective value of DCOPF), if DCOPF converge.\\
$\bullet$ $-$1000, if DCOPF or ACPF diverge.\\
$\bullet$ $+$1000, an extra reward if system finally reaches a new steady-state \textbf{at the last stage}. \\
Those values ($\pm$1000) are based on by trial-and-error experiments.

2) Action design\\
$\bullet$ If the line flow limit is too low, the DCOPF might not converge due to the narrow feasible region. On the contrary, if the line flow limit is too high, the feasible region also becomes large. However, the obtained optimal solution might lead to an operation point with \textit{tighter} power flow status on each branch, resulting in cascading failures at the next stage of the MSCF problem. Thus, the “branch flow limit” $F_l^{max}$ in the previous DCOPF formulation (2) is adopted as the action in the RL learning framework. 

3) State design\\
$\bullet$ Several quantities of each bus and the power flow of each branch are chosen and packed as the state, i.e., state=$[branch\_loading\_status, V_1,\theta_1, P_1, Q_1,..., V_n, \\ \theta_n, P_n, Q_n]$, where, $branch\_loading\_status$ are the percentage values calculated by dividing each branch flow by its loading limit for all the branches; $V_i, \theta_i, P_i, Q_i (i=1…n)$ are respectively the voltage magnitude, voltage angle, active power injection, and reactive power injection of each bus.

4) Environment\\
$\bullet$ In this study, the learning environment in the RL framework is just the power grid itself. Thus, a co-simulation platform based on DIgSILENT and MATLAB is implemented. A professional tool DIgSILENT \cite{DIgSILENT} is adopted as the simulator (\textit{environment}) to provide all needed information (\textit{states} and \textit{rewards}) to the RL network for training. Besides, the concept of \textit{step} within one independent \textit{episode} corresponds to one \textit{stage} in the MSCF problem. 

\begin{figure}[ht]
\vskip 0.1in
\begin{center}
\centerline{\includegraphics[width=\columnwidth]{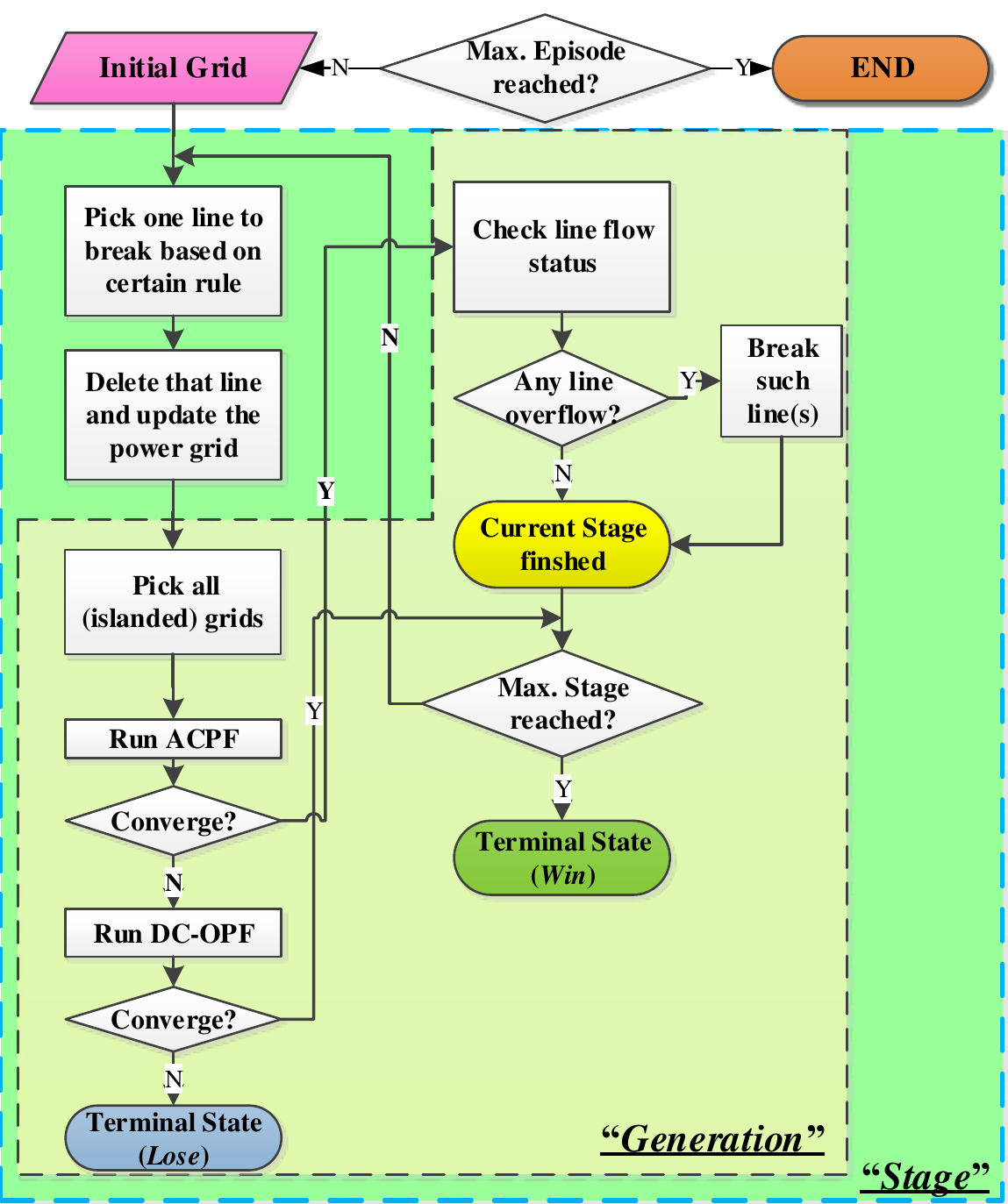}}
\caption{The overall workflow of grid simulation for MSCF study.}
\label{overall workflow}
\end{center}
\vskip -0.2in
\end{figure}

Finally, the overall workflow of the power grid simulation for the MSCF study is shown in Figure \ref{overall workflow}.

\section{Case Study}

\begin{figure}[ht]
\vskip 0.1in
\begin{center}
\centerline{\includegraphics[width=\columnwidth]{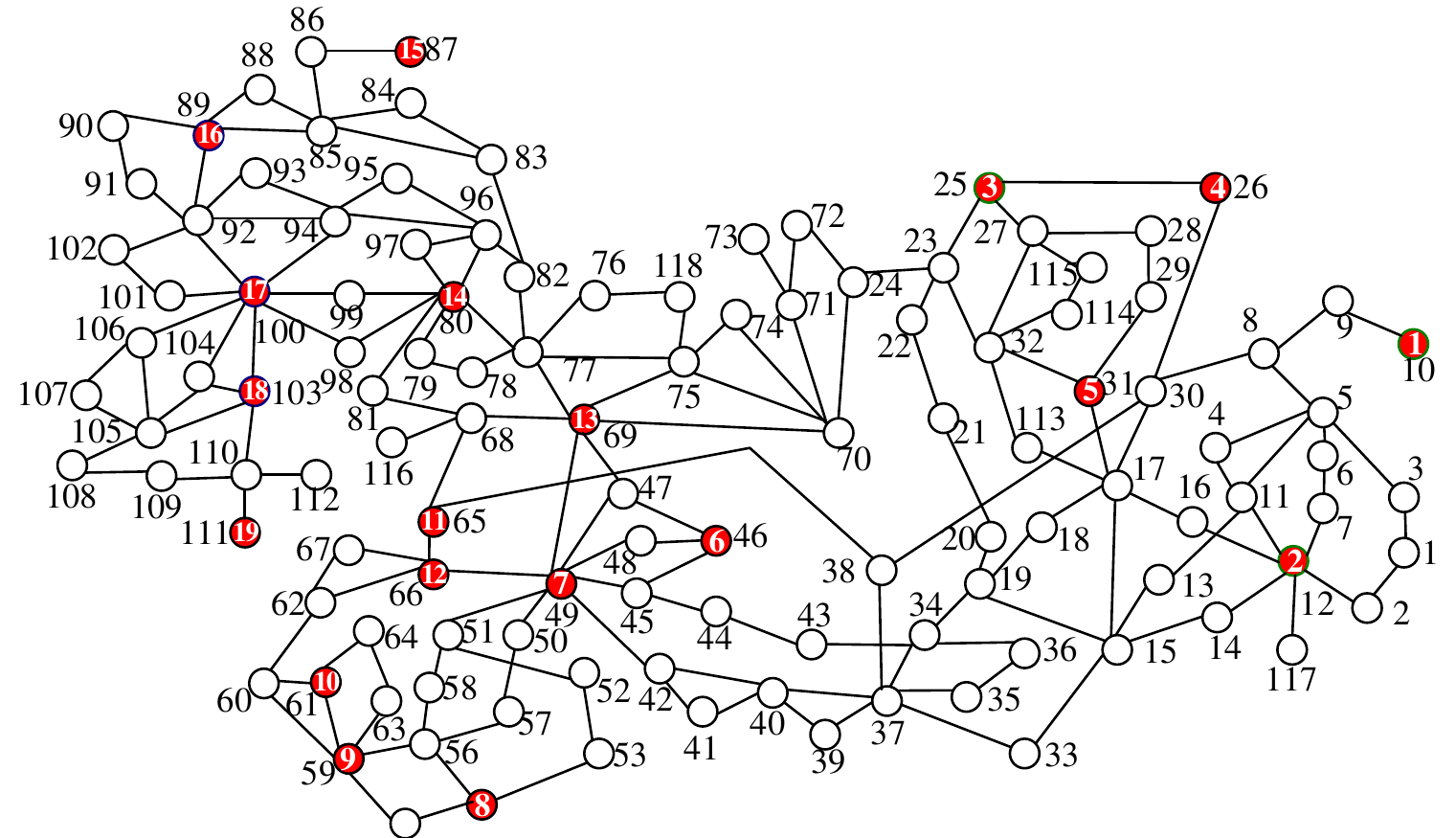}}
\caption{The topology of IEEE 118-bus power systems.}
\label{118 system}
\end{center}
\vskip -0.3in
\end{figure}

In this section, a modified IEEE 118-bus system is adopted as the test-bed for the proposed MSCF mitigation strategy. The maximum stage number is set to 3. It contains 137 buses and 177 lines (parallel lines included), 19 generators, and 91 loads. The system topology is shown in Figure \ref{118 system}, where the red dot stands for generators.

\subsection{Shallow Neural Network}

The architecture for the shallow neural network is that: one input layer, one output player,  and one hidden layer with 10 neuron units. Its input is a 1-D vector with 753 (=137$\times$4+177+28) elements; the output is the action in the RL framework (i.e., the line flow limit, c.f. Section 3).

Since both the hidden-layer dimension and output-layer dimension of the shallow network are one, the \textit{SARSA} (On-policy TD) method is employed. During the training, the \textit{action} is bounded by the range [0.80, 1.25].

\subsection{Deep Neural Network}
1) Feature engineering

To create an image-like input for the convolutional layer, the length of the original input (753) is extended to 784 = 28$\times$28 by appending extra zeros.

\begin{figure}[ht]
\vskip 0.1in
\begin{center}
\centerline{\includegraphics[width=\columnwidth,keepaspectratio]{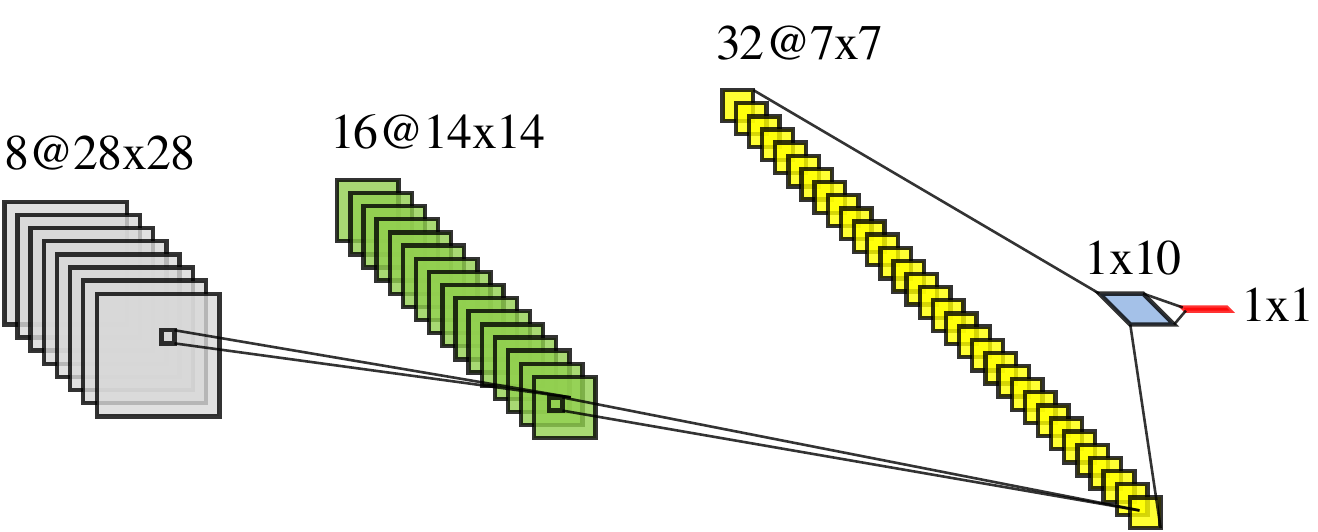}}
\caption{The network structure used in Deep RL.}
\label{DRL network}
\end{center}
\vskip -0.2in
\end{figure}

2) Network structure
Typically, a deeper network and more layers might lead to over-fitting in practice. Thus, the network structure used in this paper is shown in Figure \ref{DRL network}.

3) The \textit{Q}-learning (Off-policy TD) method is applied on it. The output of the 2nd-last layer (dimension 1 $\times$ 10) will be used in both  $ \epsilon -greedy$  policy and  $greedy$ policy. The last-layer output (dimension 1$\times$1) will be finally used to update the \textit{Q}-network parameters. The candidate set of \textit{action} is [0.8, 0.85, 0.9, 0.95, 1.0, 1.05, 1.1, 1.15, 1.20, 1.25] $ \in  \textbf{R}^{10}$.

\subsection{Experiments and Results}

\begin{figure}[ht]
\vskip 0.1in
\begin{center}
\centerline{\includegraphics[width=\columnwidth, keepaspectratio]{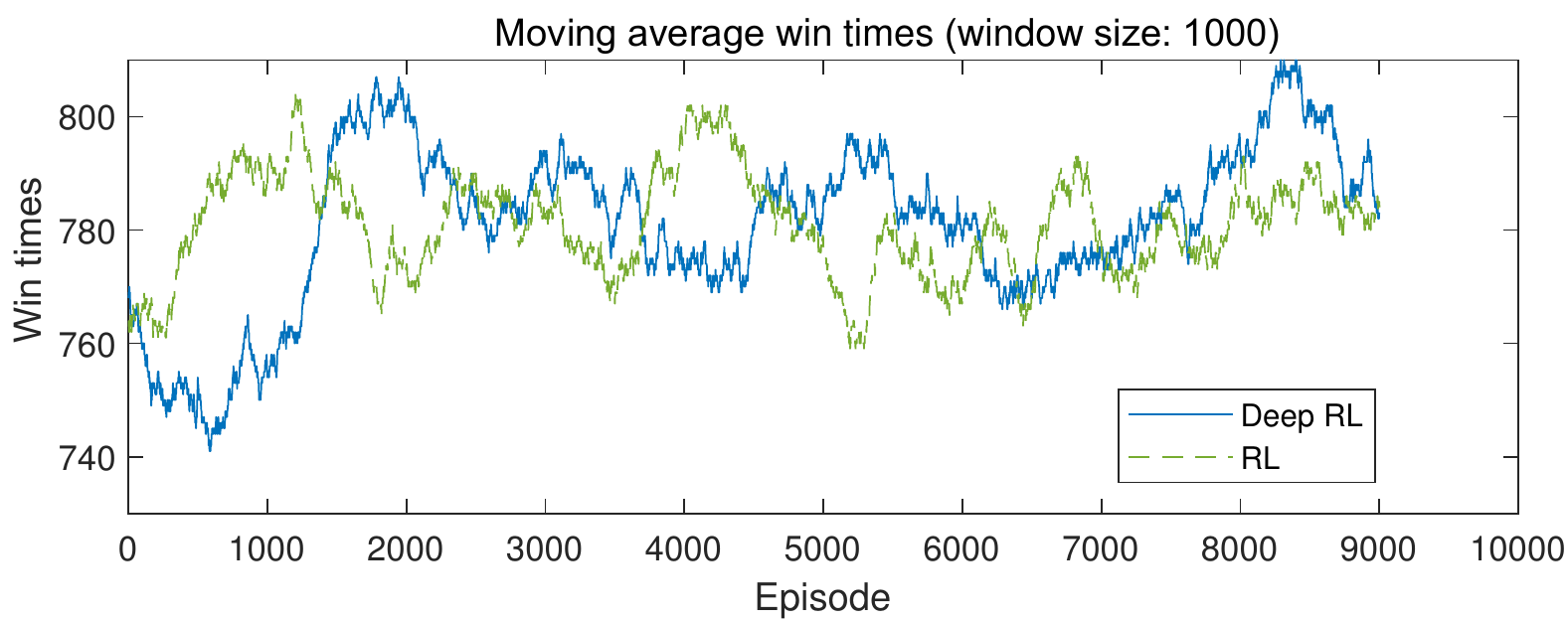}}
\caption{The learning performance by the DRL (10000 episodes) (best viewed in color).}
\label{icmlfig4_2021}
\end{center}
\vskip -0.1in
\end{figure}

For both networks, the maximum episode number is 10000, the learning rate is $10^{-4}$, and the discount rate $\gamma$ is 0.7. The learning performance is shown in Table \ref{table3}. The plot of moving average \textit{reward} (window size = 1000) for deep network case is shown in Figure \ref{icmlfig4_2021}. It can be observed that both RL and Deep RL have achieved satisfactory results in terms of winning rates (i.e., lower cascading risks). In both cases, the average return per episode is more than half of the maximum possible value (i.e., 500 = 1000/2), which shows a positive learning ability of the RL agent in mitigating cascading failures.

\begin{table}[t]
\caption{Learning Performance.}
\label{table3}
\vskip 0.15in
\begin{center}
\begin{small}
\begin{sc}
\begin{tabular}{lcc}
\toprule
Performance & \makecell{Shallow \\ Network}& \makecell{Deep \\ Network} \\
\midrule
\textit{Winning rate}    & $78.00\%$    & $78.07\%$    \\
\textit{Avg. reward}   & 640.08    & 630.46    \\
\bottomrule
\end{tabular}
\end{sc}
\end{small}
\end{center}
\vskip -0.1in
\end{table}

\section{Conclusions}
In this paper, a reinforcement learning-based mitigation strategy for the Multi-Stage Cascading Failure problem is proposed. The trained RL agent works effectively on the IEEE 118-bus system under both shallow and deep architectures with an approximately 78\% chance to avoid the power grid collapse. Potential benefits of the proposed idea in this paper include 1) enhanced resilience to extreme weather events and 2) increased penetration level of renewable sources. Investigating the effects of hyper-parameters (e.g., layer numbers, hidden neuron units, learning rate, reward amount, discount factor) of the RL network on the mitigation performance will be the next step.



\bibliography{example_paper}
\bibliographystyle{icml2019}

\end{document}